\documentclass[letterpaper]{article} % DO NOT CHANGE THIS
\usepackage{aaai23}  % DO NOT CHANGE THIS
\usepackage{times}  % DO NOT CHANGE THIS
\usepackage{helvet}  % DO NOT CHANGE THIS
\usepackage{courier}  % DO NOT CHANGE THIS
\usepackage[hyphens]{url}  % DO NOT CHANGE THIS
\usepackage{graphicx} % DO NOT CHANGE THIS
\usepackage{natbib}  % DO NOT CHANGE THIS AND DO NOT ADD ANY OPTIONS TO IT
\usepackage{caption} % DO NOT CHANGE THIS AND DO NOT ADD ANY OPTIONS TO IT
\usepackage{algorithm}
\usepackage{algorithmic}

\usepackage{newfloat}
\usepackage{listings}
\DeclareCaptionStyle{ruled}{labelfont=normalfont,labelsep=colon,strut=off} % DO NOT CHANGE THIS
\floatstyle{ruled}
\newfloat{listing}{tb}{lst}{}
\floatname{listing}{Listing}
\usepackage{amsmath}
\usepackage{amssymb}
\usepackage{booktabs}
\usepackage{multirow}
\usepackage{bbding}
\usepackage{verbatim}
\usepackage{textcomp,booktabs}
\usepackage[usenames,dvipsnames]{color}
\usepackage{colortbl}
\definecolor{mygray}{gray}{.9}

\title{Language-Assisted 3D Feature Learning for Semantic Scene Understanding}
\author{
    Junbo Zhang\textsuperscript{\rm 1}
    Guofan Fan\textsuperscript{\rm 2}
    Guanghan Wang\textsuperscript{\rm 1} 
    Zhengyuan Su\textsuperscript{\rm 1}
    Kaisheng Ma\textsuperscript{\rm 1$\ast$}
    Li Yi\textsuperscript{\rm 1,3,4}\thanks{Corresponding Author.}
}
\affiliations{
    \textsuperscript{\rm 1}Tsinghua University
    \textsuperscript{\rm 2}Xi'an Jiaotong University \\
    \textsuperscript{\rm 3}Shanghai Artificial Intelligence Laboratory 
    \textsuperscript{\rm 4}Shanghai Qi Zhi Institute  \\
    zhangjb21@mails.tsinghua.edu.cn, kaisheng@mail.tsinghua.edu.cn, ericyi@mail.tsinghua.edu.cn
}

\usepackage{bibentry}
\begin{document}

\maketitle

\begin{abstract}
Learning descriptive 3D features is crucial for understanding 3D scenes with diverse objects and complex structures.
However, it is usually unknown whether important geometric attributes and scene context obtain enough emphasis in an end-to-end trained 3D scene understanding network.
To guide 3D feature learning toward important geometric attributes and scene context, we explore the help of textual scene descriptions. Given some free-form descriptions paired with 3D scenes, we extract the knowledge regarding the object relationships and object attributes. We then inject the knowledge to 3D feature learning through three classification-based auxiliary tasks.
This language-assisted training can be combined with modern object detection and instance segmentation methods to promote 3D semantic scene understanding, especially in a label-deficient regime. Moreover, the 3D feature learned with language assistance is better aligned with the language features, which can benefit various 3D-language multimodal tasks. Experiments on several benchmarks of 3D-only and 3D-language tasks demonstrate the effectiveness of our language-assisted 3D feature learning. Code is available at https://github.com/Asterisci/Language-Assisted-3D.
\end{abstract}

\section{Introduction}

Computer vision community has studied 3D semantic scene understanding for a long time. 
Remarkable progress has been made in detecting 3D objects~\cite{Qi2019DeepHV,Xie2020MLCVNetMC} or segmenting 3D instances~\cite{Jiang2020PointGroupDP,Vu2022SoftGroupF3,Chen2021HierarchicalAF}. Most of these works struggle for learning effective 3D features to depict geometric attributes of objects as well as scene context. Though these features are commonly believed to be crucial for semantic scene understanding, it is not clear whether a network is truly emphasizing these features while being end-to-end trained for a specific 3D task.

\begin{figure}[t]
\begin{center}
\centerline{\includegraphics[width=0.98\columnwidth]{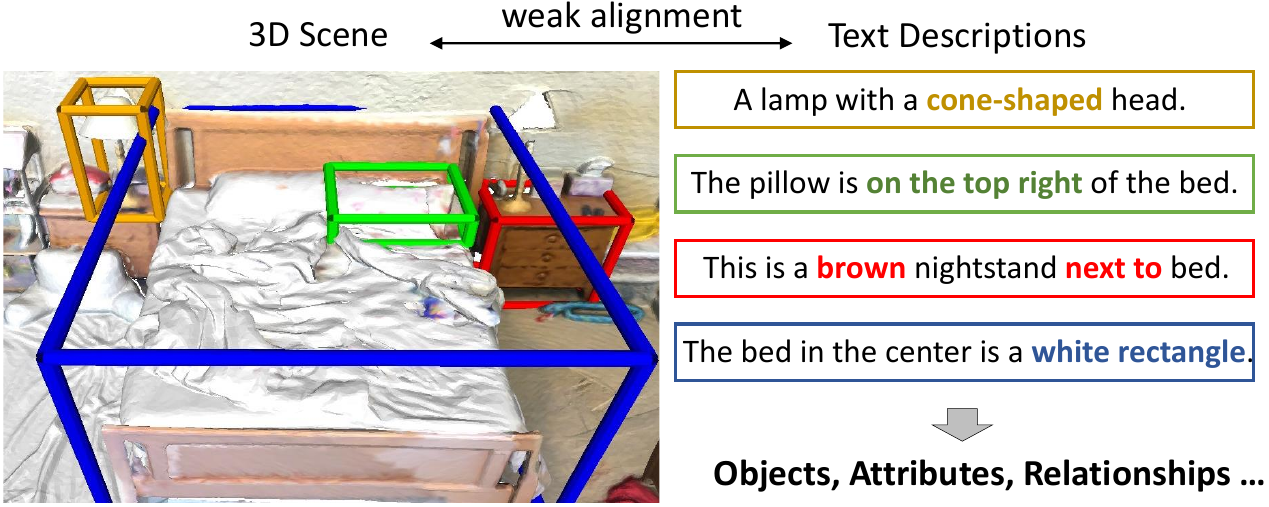}}
\caption{Text-scene pairs used in language-assisted 3D feature learning. “Weak alignment” means that text-object alignment annotations are not needed during training.}
\label{front}
\end{center}
%\vskip -0.1in
\vspace{-0.9cm}
\end{figure}

Our goal is to improve the learning process so that these important 3D features can be explicitly emphasized. And our idea comes from the following observation: humans could concisely summarize the key geometric attributes of an object as well as its surrounding context with a short natural language description (see Figure~\ref{front}) and these descriptions are very informative for identifying the object of interest. We then ask the question: can free-form languages guide 3D feature learning? We aim to imbue 3D feature learning with weak supervision knowledge from language descriptions.

3D scene understanding should benefit from text descriptions in both context modeling and object attribute learning. First, object relationships and context information in the language help understand the structure of the scene and localize specific objects. For example, “\textit{the towel hanging on a rack}” and “\textit{the chair facing the door}” provide spatial position and pose information of the objects, which could guide the scene understanding process when the \textit{towel} and \textit{chair} are hard to detect. Second, the text descriptions contain rich attribute information about the objects in a scene. For example, “\textit{a brown wooden door}” and “\textit{a small circular table}” provide color, shape and size information of the objects, which help to understand their semantic and geometric nature. 

To emphasize the two types of features mentioned above, we design a language-assisted 3D feature learning paradigm to introduce the language prior into 3D features. Previous 2D-language joint representation learning methods, such as CLIP~\cite{Radford2021LearningTV}, achieve similar objectives by aligning the global features from the 2D encoder and language encoder. However, unlike 2D images, 3D scenes are much more complex and contain a richer variety of geometry and objects. One text description corresponds to only a few objects and a small part of the scene. Moreover, the amount of 3D-language data pairs is very small compared to image-text pairs mined from the web. Therefore, we propose to parse the text descriptions into scene graphs~\cite{Armeni20193DSG,Wald2020Learning3S,Zhang2021ExploitingER} with object categories, relationships and attributes, and integrate the parsed information into 3D features at object-level. We demonstrate that this is a more precise way of introducing linguistic information into 3D models with limited paired data. Notice that text data is easy to collect since text-object alignment annotations are not needed in our method.

We find the supplemented linguistic information in our feature learning paradigm especially beneficial for data-efficient 3D scene understanding~\cite{Hou2021ExploringD3}, where limited annotations of the 3D data are available during training. Given that the annotation for 3D scenes is laborious while collecting text descriptions is relatively easy, exploring the guidance from language is a promising direction for data-efficient 3D scene understanding.

Moreover, language guidance promotes the compatibility of our learned 3D features with 3D-language tasks, such as 3D visual grounding~\cite{Zhao20213DVGTransformerRM,Yuan2021InstanceReferCH,Huang2021TextGuidedGN,Roh2021LanguageReferSM}, 3D captioning~\cite{Chen2021Scan2CapCD,Wang2022SpatialityguidedTF} and 3D visual question answering~\cite{Azuma2022ScanQA3Q,Yan2021CLEVR3DCL}. Previous methods for these 3D-language tasks usually follow a two-stage process: first, use an off-the-shelf 3D detector to extract region proposals, and then, predict the required outcome with a multimodal interaction module. They mainly focus on multimodal reasoning in the second stage. Different with previous 3D-language methods, we focus on the bottleneck of 3D feature learning: the object detectors are typically trained for general object detection of close-set categories, while a text description may refer to visual entities with various attributes and contexts. Therefore, 3D visual features learned for classical 3D scene understanding tasks might be incompatible with multimodal reasoning. We demonstrate our language-assisted 3D feature learning could promote 3D-language multimodal tasks.

Our feature learning method can be easily combined with the state-of-the-art bottom-up 3D object detection and segmentation methods, provided with natural language descriptions. Notice that the text-object alignment annotations are not needed in our method, the additional language data paired with each 3D scene can be collected with minimal effort. We test our method using various backbones and models in both 3D scene understanding benchmarks and 3D-language tasks, achieving significant progress. Our contributions can be summarized as follows:

% \vspace{-0.21cm}
\begin{itemize}
\item To the best of our knowledge, this is the first work to leverage weak language supervision to promote 3D semantic scene understanding.
\item We propose a language-assisted training method for 3D scene understanding, introducing linguistic object relationship and attribute information into 3D features. Our method can be easily combined with state-of-the-art 3D detection and instance segmentation methods.
\item With several benchmark experiments of 3D-only and 3D-language tasks, we demonstrate the benefits of introducing language prior into 3D scene understanding model, especially in a label-deficient regime.
\end{itemize}

\section{Related Work}

\subsection{3D Object Detection and Instance Segmentation}

3D object detection models take 3D data, such as point cloud and voxels, as inputs and generate 3D bounding boxes to localize the objects in a scene. Most existing methods can be considered a bottom-up manner, which requires a point grouping step to obtain object features and a detection head to generate the bounding box~\cite{Pan20213DOD,Shi2020PVRCNNPF,Zhang2020H3DNet3O}. VoteNet~\cite{Qi2019DeepHV} applies Hough Voting to group the points and vote to the similar center region. MLCVNet~\cite{Xie2020MLCVNetMC} further allows VoteNet to aggregate global contextual information via self-attention. In 3D instance segmentation, this bottom-up manner and grouping mechanisms are also widely used to generate instances. OccuSeg~\cite{Han2020OccuSegO3} adopts learnt occupancy signals to guide clustering. PointGroup~\cite{Jiang2020PointGroupDP} proposes to cluster points with dual coordinate sets, which consider both the boundary and the centroid of an instance. We propose that these 3D detection and instance segmentation methods can be further promoted with additional language supervision to remedy their inherent flaws. Based on the object-level cluster features and the language describing the objects, the proposed language-assisted training method introduces object relationship and attribute information into 3D features, which helps to better recognize and localize the objects in the scene.

\begin{figure*}[t]
\begin{center}
\centerline{\includegraphics[width=2.0\columnwidth]{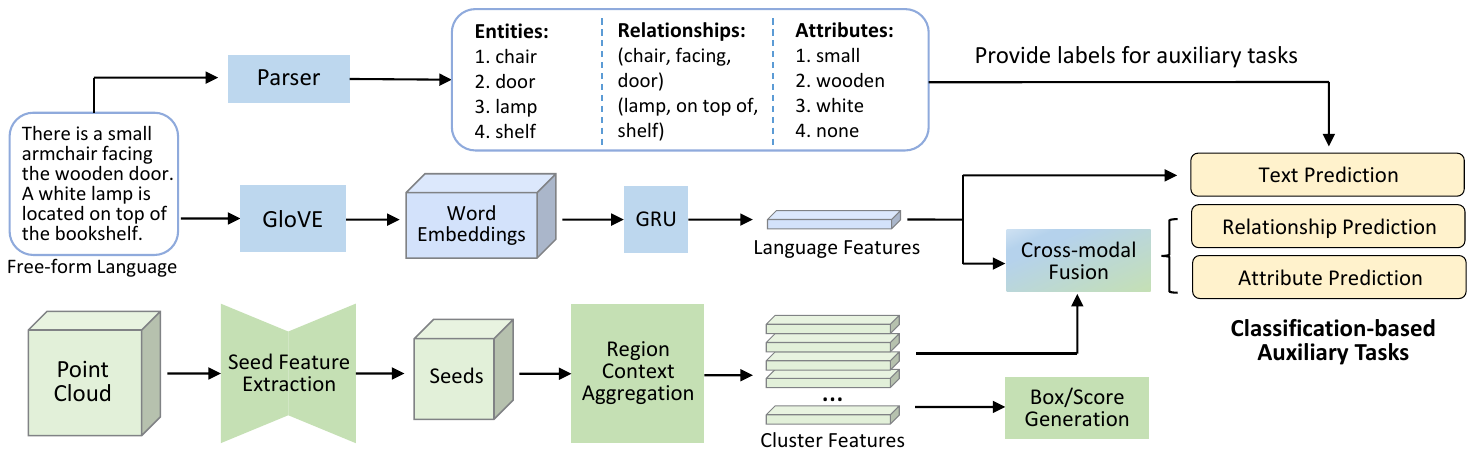}}
\caption{Overview of language-assisted training. The bottom-up 3D scene understanding branch at the bottom includes three modules: seed feature extraction, region aggregation, and box/score generation. The language processing branch at the top includes two modules: A language parser extracts entities, object relationships and attribute information from the text description. And a language model encodes the text tokens as a language feature. We fuse the language feature and object-level 3D features with a lightweight cross-attention module to perform three classification auxiliary tasks, which introduce the language prior into 3D features. The labels of these classification tasks are provided by the language parsing results.}
\label{main_fig}
\end{center}
%\vskip -0.1in
\vspace{-0.7cm}
\end{figure*}

\subsection{Multi-Modality Learning} 

\textbf{2D \& Language.} For image-language reasoning, a series of works~\cite{Lu2019ViLBERTPT,Zhang2021VinVLRV,Wang2021SimVLMSV,Desai2021VirTexLV} perform BERT-like pretraining to obtain a powerful modality interaction module. Another line of work~\cite{Jia2021ScalingUV,Zhong2022RegionCLIPRL,Li2022GroundedLP} trains the image and text encoder with contrastive learning using web-scale image-text pairs. CLIP~\cite{Radford2021LearningTV} first proposes to apply language supervision to the learning of images and achieves significant results in zero-shot learning. Different with 2D-language pre-training methods, we focus on promoting the 3D feature encoding module with weak language supervision, instead of the modality interaction module. We demonstrate that text-scene pairs help to remedy the inherent flaws in current 3D feature learning methods. Besides, compared with 2D data, 3D data contains a wider variety of geometry but a much smaller amount of data. Thus we propose to parse the texts and scenes into object-level, and introduce more accurate information (e.g. attribute \& relation) rather than just object concept information from the language.

\textbf{2D \& 3D.} Many works leverage 2D information fused with 3D data for 3D scene understanding~\cite{Ku2018Joint3P,Qi2020ImVoteNetB3,Dai20183DMVJ3,Chiang2019AUP,Hou20193DSIS3S}. The effectiveness of 2D data has been explored in 3D representation learning~\cite{Liu2021LearningF2,Li2022SimIPUS2,Afham2022CrossPointSC,Liu20213Dto2DDF}. These works utilize contrastive learning to introduce the image prior to the 3D model. 

\textbf{3D \& Language.} In 3D-language multimodal learning, shape-text joint embedding learning~\cite{Tang2021Part2WordLJ,Koo2022PartGlotLS} has been explored to facilitate the shape-text generation~\cite{Liu2022TowardsIT} and retrieval tasks~\cite{Chen2018Text2ShapeGS}. In contrast, we consider the multimodal learning of language and 3D objects in a scene for the first time, which has more complex forms of data correspondence and broader applications in 3D perception tasks.

\subsection{3D-language Tasks} 

3D vision and language tasks have recently attracted much attention. Chen et al.~\cite{Chen2018Text2ShapeGS} introduce descriptions for ShapeNet~\cite{Chang2015ShapeNetAI} objects, which enable text-to-shape generation~\cite{Sanghi2022CLIPForgeTZ,Liu2022TowardsIT} and shape captioning~\cite{Han2020ShapeCaptionerGC}. On the scene level, Chen et al.~\cite{Chen2020ScanRefer3O} release the ScanRefer dataset, which annotates natural language descriptions on ScanNet~\cite{Dai2017ScanNetR3}. ReferIt3D~\cite{Achlioptas2020ReferIt3DNL} also proposes Nr3D and Sr3D datasets for distinguishing fine-grained objects in ScanNet. These datasets facilitate the study of 3D visual grounding~\cite{Chen2020ScanRefer3O,Achlioptas2020ReferIt3DNL,Zhao20213DVGTransformerRM,Yuan2021InstanceReferCH,Chen2021D3NetAS,Abdelreheem20223DRefTransformerFO}, 3D captioning~\cite{Chen2021Scan2CapCD,Wang2022SpatialityguidedTF} and 3D question answering~\cite{Azuma2022ScanQA3Q,Yan2021CLEVR3DCL} on scene data. Current methods for 3D-language tasks mainly focus on the interaction of the two modalities for better reasoning, instead of 3D feature learning that we focus on. However, the 3D feature learning part is a bottleneck for these tasks, due to the difficulty of detecting objects in 3D point clouds and the inconsistency between visual backbones and multimodal reasoning. We focus on leveraging attribute and context information to 3D features, which can bring general improvements to 3D-language tasks.

\section{Method}
\vspace{-0.15cm}
\subsection{Overview}

Our goal is to introduce language information from paired text descriptions into 3D features. There are three critical problems in our designs. The first is what kind of knowledge from the language should be extracted. The second is how to leverage language information into 3D features without text-object alignment annotations. The third is which level of features in the model needs to be introduced with language information. For question 1, we propose to extract object relationship and attribute information using an off-the-shelf language parser. For question 2, we introduce the cross-modal knowledge by training the fused language and 3D features to perform a series of auxiliary tasks. For question 3, in the 3D models, object-level features are used to perform the auxiliary tasks, since the core subjects of a language description are usually some objects in the scene.

Figure \ref{main_fig} shows an overview of our language-assisted training architecture for 3D scene understanding. There are mainly two branches in our method: the language processing branch at the top (in blue), and the 3D scene understanding branch at the bottom (in green). 

For the 3D scene understanding branch, we ground our method on popular bottom-up object detection and instance segmentation methods, which generally include three modules as shown in Figure ~\ref{main_fig}: seed feature extraction in which a backbone processes point clouds into a dense set of features, region context aggregation that summarizes spatial regions (such as voting and grouping~\cite{Qi2019DeepHV}, RoI pooling~\cite{Shi2019PointRCNN3O} and clustering~\cite{Jiang2020PointGroupDP}) to produce a set of cluster features, and box/score generation module that outputs scene understanding results from each cluster feature. We take the cluster features after the region context aggregation for cross-modal fusion.

For the language processing branch, we take the free-form language paired with each scene as input, which is tokenized into word embeddings using a pretrained GloVE~\cite{Pennington2014GloVeGV} model. The sequences of word embedding vectors are fed into a language encoder to aggregate the textual information and extracted as a language feature. Meanwhile, the free-form text descriptions are parsed and post-processed to extract specific information, such as the entities contained in the language and their relationships and attributes.

After feature extracting and language processing, the object-level 3D features and language features are fused with a lightweight cross-attention module and used to perform three classification-based auxiliary tasks. The first is \textit{text classification} that supervises the language encoder. The remaining tasks are \textit{relationship classification} and \textit{attribute classification} which help to introduce linguistic information into 3D features. We choose the most frequent relationships and attributes from the language parsing results as the classification ground truth for each task. By training these auxiliary objectives, the 3D features become “language-aware”, and the linguistic information overlooked by traditional can be supplemented.

Different with 2D-language training methods for multi-modal downstream tasks~\cite{Kamath2021MDETRM}, we only train the text/3D feature encoder with weak text-scene alignments, instead of multi-modal reasoning module. After language-assisted training, 3D model can be used to perform 3D-only tasks, while the language encoder and 3D model can be used in 3D-language tasks. Next, we describe the language processing and the design of auxiliary tasks in detail.

\vspace{-0.15cm}

\subsection{Language Parsing and Post-processing}

\vspace{-0.15cm}

Given a free-form text, we aim to extract all the entities mentioned in the language and parse their relationships and attributes. First, we deal with the problem of pronouns’ reference. Natural languages often face the long-distance coreference problem, such as “\textit{This is a brown chair. It is located under the whiteboard.}”, where “\textit{chair}” and “\textit{it}” refer to the same 3D object in the scene. Following Feng et al.~\cite{Feng2021FreeformDG}, we replace pronouns with the noun phrases of the referred 3D object in the descriptions. Because the statistical analysis of the languages in the dataset shows that the pronouns refer to the referred 3D object in most cases.

Then, we utilize a rule-based language parser to extract the entities and their relationships in the description. We add additional rules to the language parser to adapt the model to the 3D scene data and corresponding language descriptions. After parsing the language, we obtain several noun phrases, e.g., “\textit{a small armchair}”, “\textit{the brown circle table}”, each representing an entity in the scene. Notice that the language parser combines the attribute and the noun words for each entity to form a noun phrase. In addition, we can obtain the relationships of the entities, which are expressed in the triple form: \textit{(subject, relation, object)}. For example, for the description “\textit{The small chair is facing the door}”, the resulting relationship of the two entities is \textit{(chair, facing, door)}.

Given the noun phrases and their relationships extracted from the language, we select the most frequent relation phrases and attribute words in the dataset. In total, we select 8 relationships, of which the frequency is greater than 500, and 26 attributes, of which the frequency is greater than 100. The selected attributes can be further divided into color, shape and size. Then we match entity names to the object classes in the 3D dataset. To sum up, for each text description, we find all the entities and save their relationships and attributes. These will be regarded as the language prior and used as the ground truth labels for the auxiliary tasks.

\subsection{Language-assisted Training for 3D Semantic Scene Understanding}
\label{auxiliarytask}

In order to leverage language prior to 3D models, we propose to learn three auxiliary classification tasks along with the 3D perception task. The auxiliary tasks are based on the outputs of 3D encoder, language encoder and the language parsing result. We first introduce the data preparations and upstream models for the auxiliary tasks and then the detailed design of each task.

\textbf{Preparation for the Auxiliary Tasks.} Given a 3D scene data and a paired text description, we extract global language features and object-level 3D features with a language encoder and a 3D detection/segmentation model. The word tokens from the text description are embedded using a pretrained GloVE~\cite{Pennington2014GloVeGV} model and then fed into a GRU cell~\cite{Chung2014EmpiricalEO} to aggregate the textual information. We take the final hidden state of the GRU cell as the final language embedding $f_{lang} \in \mathbb{R}^{c_l}$. The point cloud data is fed into the seed feature extraction and region context aggregation modules in the 3D scene understanding model to obtain the object-level clusters $P=\{x_i,f_i\}_{i=1}^{M}$, where $M$ is the number of clusters. $x_i \in \mathbb{R}^3$ and $f_i \in \mathbb{R}^{c_v}$ are cluster center and feature, respectively. $c_l$ and $c_v$ are the channel numbers of the language and cluster features. Next, the cluster features are fed to the following module to output the 3D scene understanding results, such as box regression in VoteNet ~\cite{Qi2019DeepHV} and score prediction in PointGroup~\cite{Jiang2020PointGroupDP}. Meanwhile, a semantic classification score $s_i \in \mathbb{R}^{N_c} $ is predicted for each cluster, where $N_c$ is the number of object classes in the 3D dataset.

The extracted language feature and object-level 3D features are then fed into a lightweight cross-attention module to integrate multimodal information. A QKV attention module is used, where language embedding $f_{lang} \in \mathbb{R}^{c_l}$ serve as query vectors and 3D features $f_i \in \mathbb{R}^{c_v}$ in the point cluster $P$ as key/value vectors to compute the final multi-modal features. The fused multimodal features, the semantic classification scores $S=\{s_i\}_{i=1}^{M}$ and the language parsing results are used to build the auxiliary tasks as followings.

\textbf{Text Classification.} To supervise the language encoder, we include an object classification loss based on the input description as in ScanRefer~\cite{Chen2020ScanRefer3O}. We consider the benchmark classes in the corresponding 3D dataset and apply a linear classifier to the language embedding. The language-to-object classification loss $L_{text}$ is a multi-class cross-entropy loss.

\textbf{Relationship Classification.} If any relationship triplet $t=(s,r,o)$ is extracted from the text description, we perform a relation classification task, aiming to enhance the contextual information and the awareness of object relationships for each object feature. $s,r,o$ represent subject, relationship and object class, respectively. For each triplet, we first select all the fused multimodal features that the predicted semantic class $c_i$ matches the subject ($s$) and object ($o$) class. The indexes of the selected features for subjects and objects are denoted as $I=\{i:c_i=s\}$, $J=\{j:c_j=o\}$, $1\le i,j\le M$, respectively. If using detection backbones, we only choose the cluster features that the predicted objectness mask $m_i \in \{0, 1\}$ is equal to 1, indicating that the box regressed from this feature encloses an object. Assuming $N_{sub}$ subject features and $N_{obj}$ object features are selected, we then concatenate each subject and object feature and obtain $N_{sub} \times N_{obj}$ pairs of features for relationship classification.

Notice that the relationship classes we considered are not mutually exclusive.
% We assume that, realistically, for a specific subject-object pair, there are multiple valid relationships that describe their interaction.
For instance, a chair “\textit{next to}” a door could also be “\textit{facing}” the door.
Therefore, we formulate the relationship prediction loss $L_{rel}$ as per-class binary cross-entropy. This way, whether a subject-object pair should be assigned to a specific relationship label is judged independently. In practice, for each triplet, we have only one ground truth relationship type. Thus, we use the dependencies between relationships to build labels for the other relationships. For instance, if the triplet \textit{(chair, left of, table)} is given, the relationships between the two mentioned objects can be determined to be “\textit{left}” and “\textit{next to}”, and not to be “\textit{above}” or “\textit{under}”. However, we can not determine whether the relationship between two objects is “\textit{facing}”. So, in this case, the labels used in the per-class binary cross-entropy are $1$ for “\textit{left}” and “\textit{next to}”, $0$ for “\textit{above}” and “\textit{under}”, and $0.5$ for “\textit{facing}”. The label of $0.5$ indicates that the ground truth of this category can not be determined based on the given relationship triplet.

In this way, for each relationship triplet $t=(s,r,o)$, we have $N_{sub} \times N_{obj}$ loss values:
$l_{ij} = L_{BCE}( p_{ij},\text{ } y_r), i\in I, j\in J$, where $p_{ij}$ is the relation prediction and $y_r$ is the generated label for per-class binary cross-entropy loss. Since the object features selected according to the predicted semantic class do not necessarily match the exact objects described by language, we only consider the most confident prediction and take the minimum of these loss values as the final relationship prediction loss: $L_t = \min(l_{ij})$. The overall relationship prediction loss is $\text{ } L_{rel} = \sum_{t=1}^T L_t$, where $T$ is the number of relationship triplets extracted from the text description.
% L_t = l_{i^*j^*}, \text{ }\text{ }\text{ }(i^*,j^*) = \mathop{\arg\min}\limits_{i\inI,j\in J}(l_{ij})

\begin{table}[t]
\footnotesize
\setlength\tabcolsep{2.5pt}
\renewcommand{\arraystretch}{0.9}
 \centering
  \begin{tabular}{lc|lc}
    \toprule
    Method &  mAP@0.5 & Method &  mAP@0.5 \\
    \midrule
    VoteNet & 31.67  & MLCVNet & 30.18 \\
    +Ours & \textbf{33.41} & +Ours & \textbf{33.69}  \\
    \bottomrule
  \end{tabular}
  \vspace{-0.2cm}
{
 \caption{3D object detection results on ScanNetV2.}
 \label{detection}
}
\end{table}

\begin{table}[t]
\footnotesize
  \centering
  \begin{tabular}{lccc}
  \toprule
    No. of Boxes  & 1 & 4  & All    \\
    \midrule
    VoteNet &  13.74  &  23.08   &  51.92  \\
    +Ours   &  \textbf{18.55}   & \textbf{30.61}  &  \textbf{52.60} \\
    % \midrule
    % MLCVNet &    &     &    \\
    % +Ours   &  \textbf{}   & \textbf{26.90}  &  \textbf{} \\
    \bottomrule
  \end{tabular}
  \caption{3D object detection results on ScanNetV2 using limited bounding box annotations. Each scene contains an average of about 13 bounding boxes. The metric is mAP@0.25.}
  \label{data-efficient}
 \vspace{-0.5cm}
\end{table}

\begin{table}[t]
\footnotesize
\setlength\tabcolsep{2.5pt}
\renewcommand{\arraystretch}{0.9}
\centering
 \begin{tabular}{lcc}
    \toprule
    Method  & mAP@0.25 & mAP@0.5 \\
    \midrule
    PointGroup   & 71.3 & 55.7    \\
    +Ours  & \textbf{72.3}  & \textbf{56.9}   \\
    \bottomrule
  \end{tabular}
{
 \caption{3D instance segmentation results.}
 \label{segmentation}
}
\vspace{-0.4cm}
\end{table}

\textbf{Attribute Classification.} For each entity in the text description, we select the corresponding multimodal features according to the predicted semantic class in the same way as the relationship prediction task. A linear classifier is used to predict the attribute category for each selected feature. The classification labels are provided by the language parsing results. We group the attribute types that are not in the chosen lists into “\textit{Others}”. Similarly, we take the minimum of the loss values as the final attribute prediction loss $L_{attr}$, which is a multi-class cross-entropy loss.

\textbf{Overall Loss Function.} The final loss is a combination of the perception loss and auxiliary losses: 

\vspace{-0.37cm}
\begin{equation}
L_{overall} =  L_{perception} + \alpha L_{text} + \beta L_{rel} + \gamma L_{attr}, 
\end{equation}

where $\alpha$, $\beta$ and $\gamma$ are the weights for the individual loss terms. The perception loss $L_{perception}$ is the same as the used 3D detection/segmentation model.

\section{Experiments}
\label{exp}

\subsection{Experimental Setup}

\textbf{Dataset.} We adopt the ScanNetV2~\cite{Dai2017ScanNetR3} dataset for 3D detection and instance segmentation. ScanNetV2 provides $1,513$ indoor scans with semantic and instance segmentation annotations for $18$ object categories. Since ScanNetV2 does not provide oriented bounding box annotation, we train the detection model to predict axis-aligned bounding boxes as in VoteNet~\cite{Qi2019DeepHV}.

We perform language-assisted training based on a widely used point cloud based visual grounding dataset, ScanRefer~\cite{Chen2020ScanRefer3O}. Text-object alignment annotations are not used in our 3D feature learning process. The ScanRefer dataset provides $51,583$ free-form text descriptions based on the $800$ ScanNet scenes. Each scene has an average of $13.81$ objects and $64.48$ descriptions. 
% For language-assisted training, we only train the 3D model on the ScanNet scenes that are contained in the ScanRefer dataset and evaluate on the whole ScanNet testing set. Specifically, we train the 3D model on 562 scenes from ScanRefer's training set. 
The 3D visual grounding and 3D captioning tasks are performed on ScanRefer. We follow the ScanRefer benchmark to split the train/val/test set with $36,655$,~ $9,508$, and $5,410$ samples, respectively. Results and analysis are conducted on
the val split, as the hidden test set is not officially available.

% The 3D visual grounding task is performed on ScanRefer~\cite{Chen2020ScanRefer3O}. We follow the ScanRefer benchmark to split the train/val/test set with $36,655$,~ $9,508$, and $5,410$ samples, respectively. We use Acc@0.25IoU and Acc@0.5IoU as the visual grounding metrics, i.e., the percentage of the correctly predicted bounding boxes whose IoUs with the ground-truth bounding boxes are larger than $0.25$ and $0.5$. The overall accuracy and the accuracies on both “unique” and “multiple” subsets are reported. \textit{Unique} means there is only one object of its class in the scene. \textit{Multiple} means there are distractors of the same class as the referenced object. Results on ScanRefer validation set are reported.

% The 3D captioning task is also performed on the ScanRefer dataset using the same benchmark split. We use the same metrics as in Scan2Cap~\cite{Chen2021Scan2CapCD} to jointly measure the quality of the generated descriptions and the detected bounding boxes. Specifically, we evaluate the descriptions by combining captioning metric IoU scores between predicted bounding boxes and the target bounding boxes.

\textbf{Implementation Details.} For the detection and instance segmentation backbone, we follow the same implementation details as in VoteNet~\cite{Qi2019DeepHV}, MLCVNet~\cite{Xie2020MLCVNetMC} and PointGroup~\cite{Jiang2020PointGroupDP}. In VoteNet and MLCVNet, we obtain the object-level features from the cluster features after voting and grouping, which will be used for the following auxiliary tasks. In PointGroup, we take the cluster features after ScoreNet as the encoded object-level features. For language-assisted training, we jointly train the 3D perception task and the auxiliary tasks for 70 epochs with an Adam optimizer and an initial learning rate of 0.001. The learning rate is multiplied by 0.3 after 50 and 60 epochs. We set $\alpha$, $\beta$ and $\gamma$ to 0.1, 0.05 and 0.05. VoteNet and MLCVNet are trained with the batch size of 12, in which each scene is paired with one description. PointGroup is trained with the batch size of 4, in which each scene is paired with 8 descriptions to balance the training time. Note that each scene is trained for multiple times in an epoch since an average of 64.48 descriptions are paired with each scene. For 3D-language tasks, we train the detection backbone and the multimodal interaction module end-to-end following ScanRefer~\cite{Chen2020ScanRefer3O} and Scan2Cap~\cite{Chen2021Scan2CapCD}. All the evaluation results are the average of 5 runs. Details are contained in the appendix.

\textbf{Language Parser.} After parsing the language, we select 8 relationships and 26 attributes to form the training set of auxiliary tasks. In total, we extract relationships from $12,325$ descriptions and extract attributes from $11,402$ descriptions among the $36,665$ descriptions in ScanRefer’s training set. Our improved language parser successfully parses 83.28\% of the training texts. More language parsing results and analysis can be found in the appendix.

\begin{table}[t]
\footnotesize
\renewcommand\arraystretch{1.0}
  \centering
  \begin{tabular}{lc|ccc}
    \toprule
    Method & Input &  Unique & Multiple & Overall \\
    % & Acc@0.5  & Acc@0.5 & Acc@0.5   \\     
    % \midrule
    % \multicolumn{8}{c}{Results on ScanRefer validation set} \\
    \midrule
    ScanRefer & 3D & 47.86  & 20.50 & 25.81 \\
    +Ours  & 3D & \textbf{49.80} & \textbf{22.75} & \textbf{28.00} \\
    \midrule
    ScanRefer & 2D+3D  & 56.01  & 23.52 & 29.83\\
    +Ours & 2D+3D & \textbf{59.31} & \textbf{25.07}  & \textbf{31.71}  \\
    % \midrule
    % 3DVG & 3D &   &  &  \\
    % +Ours  & 3D & \textbf{} & \textbf{} & \textbf{} \\
    \bottomrule
  \end{tabular}
  \caption{Visual grounding results on ScanRefer. \textit{Unique} means there is only one object of its class. \textit{Multiple} means there are distractors of the same class as the referenced one.} % as the referenced object.
  \vspace{-0.1cm}
\label{scanrefer}
\end{table}

\begin{table}[t]
\footnotesize
  \centering
  \begin{tabular}{l|cc}
    \toprule
    Method &  CiDEr@0.25IoU  & CiDEr@0.5IoU \\
    \midrule
    Scan2Cap &  50.95 & 37.29   \\
    +Ours &  \textbf{52.97} & \textbf{39.06} \\
    \bottomrule
  \end{tabular}
  \caption{Captioning results on ScanRefer. We average the captioning metric with the percentage of the predicted bounding boxes whose IoU are greater than 0.25 and 0.5.} % whose IoU with the ground truth
  \label{caption}
  \vspace{-0.1cm}
\end{table}

\subsection{Object Detection and Instance Segmentation}

We first show the 3D detection and instance segmentation results on ScanNetV2 in Table \ref{detection} and Table \ref{segmentation}. VoteNet~\cite{Qi2019DeepHV} and MLCVNet~\cite{Xie2020MLCVNetMC} are used as the detection backbone, and PointGroup~\cite{Jiang2020PointGroupDP} is used as the instance segmentation backbone. We reproduce all the baseline methods and apply the language-assisted training in our method. For language-assisted training, we only train the 3D model on the ScanNet scenes that are contained in the ScanRefer dataset and evaluate the model on the whole ScanNet test set. Specifically, we train the 3D model on 562 scenes from ScanRefer's training set. The results show that the language prior from text descriptions is successfully introduced to 3D model and is beneficial for detection and instance segmentation tasks with different backbones.

\subsection{Data-Efficient 3D Scene Understanding}

We study the data-efficient 3D scene understanding to further explore the effect of language supervision. A widely used benchmark is to use limited bounding box annotations to perform 3D object detection. Following ~\cite{Hou2021ExploringD3},  we random sample \{1, 4\} bounding boxes per scene and train the detector while the average number of bounding boxes per scene is about 13. The results are shown in Table \ref{data-efficient} with the metric of mAP@0.25. It is observed that our method outperforms the vanilla method significantly when limited bounding box annotations are provided, indicating that the knowledge from language helps 3D models deduce other object and geometry information in a scene from limited annotations. Since collecting free-form descriptions paired with 3D scenes is relatively convenient, exploring the effect of language supervision is a promising direction to promote data-efficient 3D scene understanding.

\subsection{Vision-Language Task}

Table \ref{scanrefer} shows the 3D visual grounding results with ScanRefer~\cite{Chen2020ScanRefer3O} baseline on ScanRefer dataset. In the \textit{Input} column, \textit{3D} stands for \textit{xyz + RGB}, and \textit{2D + 3D} means an extra 128-dimensional multiview feature for each point is added, as indicated by ScanRefer~\cite{Chen2020ScanRefer3O}. We use Acc@0.5IoU as the visual grounding metrics, i.e., the percentage of the correctly predicted bounding boxes whose IoUs with the ground-truth bounding boxes are larger than $0.5$. The overall accuracy and the accuracies on both “unique” and “multiple” subsets are reported. \textit{Unique} means there is only one object of its class in the scene. \textit{Multiple} means there are distractors of the same class as the referenced object. Results on ScanRefer validation set are reported. The VoteNet backbone in ScanRefer architecture is trained with language assistance in our method, while the multimodal interaction module is kept the same.

Table \ref{caption} shows the 3D captioning results with Scan2Cap~\cite{Chen2021Scan2CapCD} baseline on ScanRefer dataset. The VoteNet backbone in Scan2Cap is trained with language assistance in our method. We use the same metrics as in Scan2Cap~\cite{Chen2021Scan2CapCD} to jointly measure the quality of the generated descriptions and the detected bounding boxes. We use the standard captioning metric of CiDEr~\cite{Vedantam2015CIDErCI}, and employ CiDEr@0.25IoU and CiDEr@0.5IoU where only the prediction whose IoU is larger than 0.25 and 0.5 will be considered. Specifically, we evaluate the descriptions by combining the captioning metric with IoU scores between predicted bounding boxes and the target bounding boxes. The results show that our method outperforms the baselines significantly, indicating that introducing language prior to 3D features is beneficial for general 3D-language tasks.

% \begin{table*}[t]
% \footnotesize
%   \centering
%   \begin{tabular}{cccccccc}
%     \toprule
%       &  & \textit{Cabinet}  &  \textit{Door} & \textit{Chair} & \textit{Desk} & \textit{Sink} & \textit{Shelf}  \\ 
%     \midrule
%     \rowcolor{mygray}
%      & $\Delta$mAP@0.25 & +3.03 & +2.07  & +0.34 & +1.38 & +0.03 & -0.02 \\   \rowcolor{mygray}
%     \multirow{-2}{*}{Refer Categories} & $\Delta$mAP@0.5 & +2.68 & +3.45 & +2.06 & +2.11 & +2.38 & +1.81 \\ 
%     \midrule
%     \multirow{2}{*}{Rest Categories} & $\Delta$mAP@0.25 
%      & +0.66 & +0.31  & +0.44 & +0.69 & +0.12 & +0.36 \\
%      & $\Delta$mAP@0.5 
%      & +1.21 & +1.26  & +1.24 & +0.96 & +0.72 & +0.39 \\
     
%     \bottomrule
%   \end{tabular}
%   \caption{mAP changes of specific categories (paired with more than 500 text descriptions) after applying the language assistance.}
%   \label{specific}
%   \vspace{-0.2cm}
% \end{table*}

\begin{table*}[t]
\footnotesize
  \centering
  \begin{tabular}{cccccccc}
    \toprule
      & Metric & \textit{Cabinet}  &  \textit{Door} & \textit{Chair} & \textit{Desk} & \textit{Sink} & \textit{Shelf}  \\ 
    \midrule
   \rowcolor{mygray}
    Refer Categories & $\Delta$mAP@0.5 & +2.68 & +3.45 & +2.06 & +2.11 & +2.38 & +1.81 \\ 
    \midrule
    Rest Categories 
     & $\Delta$mAP@0.5 
     & +1.21 & +1.26  & +1.24 & +0.96 & +0.72 & +0.39 \\
     
    \bottomrule
  \end{tabular}
  \caption{mAP changes of specific categories (paired with more than 500 text descriptions) after applying the language assistance.}
  \label{specific}
  \vspace{-0.4cm}
\end{table*}

\begin{table}[t]
\footnotesize
\renewcommand{\arraystretch}{1}
 \centering
  \begin{tabular}{lcc}
    \toprule
    Method &  Obj. Det. & Inst. Seg. \\
    \midrule
    Baseline  &  31.67  & 55.7 \\
    +relation  &  32.47  &  56.5 \\
    +attribute &  32.80  &  56.1 \\
    +relation+attribute  &  \textbf{33.41}  &  \textbf{56.9} \\
    \bottomrule
  \end{tabular}
  \vspace{-0.2cm}
{
 \caption{Ablation study on auxiliary tasks in object detection and instance segmentation (metric is mAP@0.5).}
 \label{ablation_aux}
}
\end{table}

\begin{table}[t]
\footnotesize
\renewcommand{\arraystretch}{1}
\centering
 \begin{tabular}{c|>{\columncolor{mygray}}cc|>{\columncolor{mygray}}cc}
    \toprule
    Text Percent & \textit{Cabinet} & \textit{Rest} & \textit{Door} & \textit{Rest} \\
    \midrule
    30\%  &  +0.31  &  +0.49 & +0.21 & +0.95  \\
    60\%  &  +1.43  &  +0.46 & +1.42 & +0.71  \\
    100\% &  \textbf{+2.68}  &  \textbf{+1.21} & \textbf{+3.45}
    & \textbf{+1.26} \\
    \bottomrule
  \end{tabular}
{
 \caption{Ablation study on the number of texts (mAP@0.5). We show results for the two categories with the most textual descriptions. This result shows the potential to further improve performance by collecting more language data.}
 \label{ablation_lang}
}
\vspace{-0.5cm}
\end{table}

\vspace{-0.2cm}
\section{Discussion and Ablation Study}
\subsection{Language-assisted Training for a Specific Category}

To further explore the effect of language-assisted training, we train the VoteNet with text descriptions only referring to a specific category. Table \ref{specific} shows the changes ($\Delta$) in object detection results of the referred category and the rest categories after applying the language assistance. We show the results of those object categories that have more than 500 text descriptions and do not belong to “\textit{Others}” class in ScanNetV2. The results show a significant boost in the detection results for the referred category, which is larger than the general improvement in Table \ref{detection}, possibly because the chosen languages contain less noise for the referred category. Meanwhile, changes of the detection results for the rest categories are relatively small as shown in the last row in Table \ref{specific}. Since the chosen descriptions describing a specific category can also involve other category objects, the detection of the rest categories can be improved to some degree. This experiment provides the insight that, in practical applications, we can use the descriptions referring to a specific category to improve the accuracy of those categories that are important for application or difficult to detect.

\vspace{-0.2cm}
\subsection{Ablation Study}
Table \ref{ablation_aux} shows the ablation study on the relationship and attribute classification auxiliary tasks. The baseline method is VoteNet and PointGroup for object detection and instance segmentation. The metric is mAP@0.5. $relation$ and $attribute$ represent the relationship and attribute auxiliary task, respectively. It is observed that each auxiliary task is effective, and using both of them can further improve the performance. Table \ref{ablation_lang} shows the ablation study on the number of text descriptions used in language-assisted training for a specific category. We show the changes ($\Delta$) of mAP@0.5 in the object detection model trained with different percentages of texts. It is observed that the detection results improve with the increase of text numbers. Given that collecting free-form languages describing a 3D scene is convenient, larger amount of text descriptions can be used to further improve the results in the future work.

\vspace{-0.2cm}
\subsection{Qualitative Comparison}
\begin{figure}[h]
\begin{center}
\centerline{\includegraphics[width=1.0\columnwidth]{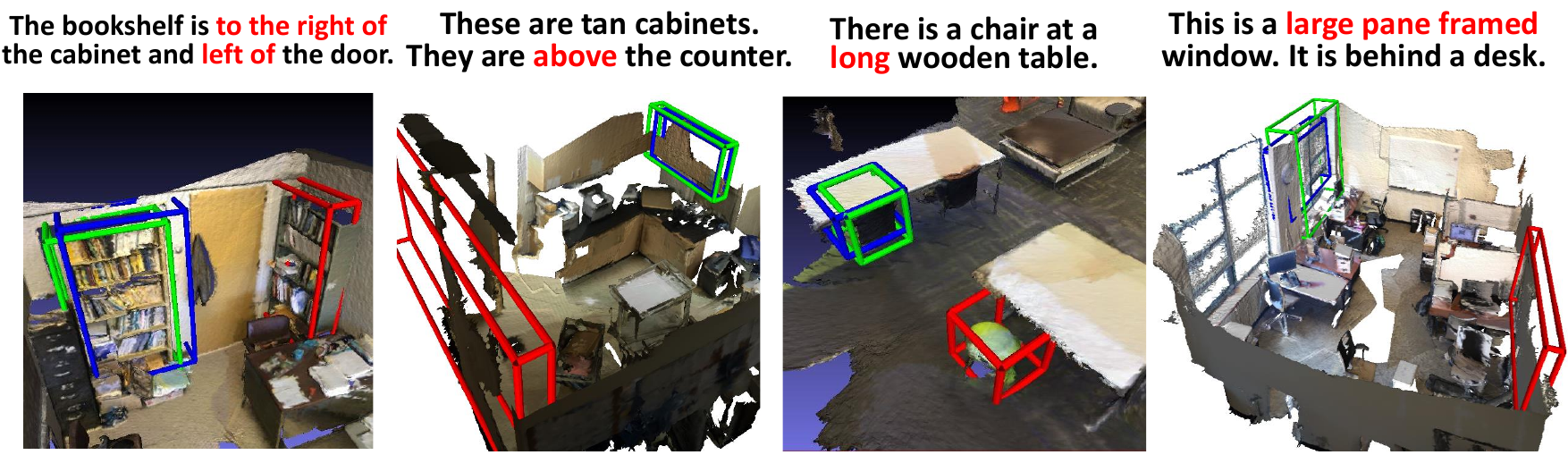}}
\caption{Qualitative comparison results on visual grounding. The ground-truth boxes are displayed in blue. The predicted boxes of our method and the baseline are displayed in green and red, respectively.}
\label{visualization}
\end{center}
%\vskip -0.1in
\vspace{-1.1cm}
\end{figure}

Figure \ref{visualization} shows the visualization of visual grounding results. The ground-truth boxes are displayed in blue. The predicted boxes of our method and the baseline are displayed in green and red, respectively. In the first picture, there are two bookshelves in the scene, and the target of the language input is the left one. Two relation triplets can be extracted from the language description, namely, \textit{(bookshelf, right of, cabinet)} and \textit{(bookshelf, left of, door)}. Our model better preserves this relation information and finds the correct bookshelf. In contrast, the baseline model fails to infer from relational information and predicts the wrong results. The second picture shows that our method utilizes the object relationship to correctly localize the \textit{cabinet} among distractors, indicating that the 3D features in our method better preserve relationship information. The failure case of ScanRefer illustrates that it can not distinguish ambiguous objects according to spatial relationships. In the third example, there are two chairs  and the target one is on the left. In the language descriptions, the attribute word \textit{“long”} used to describe the table is an essential basis for finding the target object. Our model better utilizes attribute clue and predicts the right location, while the baseline method fails to distinguish between the two tables. The last picture shows that when relationship information is insufficient to make a decision, the attribute information can be used to localize the target in our method. We provide more qualitative comparison results for both visual grounding and 3D object detection in the appendix.

\section{Conclusion}
\label{conclusion}

% This paper proposes a language-assisted training method to promote 3D scene understanding. With free-form languages paired with 3D scenes, we extract object relationship and attribute information from languages. Three auxiliary tasks are designed to transmit cross-modal knowledge based on multimodal features. We apply the proposed method to several 3D object detection and instance segmentation models. Several experiments of 3D-only and 3D-language tasks demonstrate the benefits of language supervision in 3D scene understanding. We also discuss our method’s application in data-efficient 3D scene understanding and improving the detection performance for specific categories.

This paper proposes a language-assisted training method to promote 3D scene understanding. With
free-form languages paired with 3D scenes, we extract object relationship and attribute information 
from languages and introduce this language prior to object-level 3D features. Three auxiliary
tasks are designed to transmit cross-modal knowledge based on multimodal features. We apply
the proposed method to several 3D object detection and instance segmentation models. Several
benchmark experiments of 3D-only and 3D-language tasks (e.g. visual grounding, captioning) demonstrate the benefits of language supervision in 3D scene understanding. We also discuss our method’s application in data-efficient 3D scene understanding and improving the detection performance for specific categories.

\setlength{\bibsep}{0.18ex}

\bibliography{aaai23}

\appendix \section{Supplementary Material}

\vspace{0.25cm}

\appendix \section{Implementation Details}

\subsection{Model Details}
\label{model_details}

\textbf{Cross-modal fusion module.} In the main paper, a lightweight cross-modal fusion module is used to fuse the language feature and object-level 3D features. We implement this module with a multi-head attention layer with only one head followed by layer normalization.

\textbf{Language-assisted training with modern object detection and instance segmentation models.} In VoteNet~\cite{Qi2019DeepHV} and MLCVNet~\cite{Xie2020MLCVNetMC}, we take the cluster features after vote aggregation (output of the PointNet~\cite{Qi2017PointNetDL} in the proposal module) as the object-level features. In PointGroup~\cite{Jiang2020PointGroupDP}, we take the cluster features after ScoreNet as the encoded object-level features. The ScoreNet utilizes a U-Net to aggregate the information of cluster points. And a cluster-aware max-pooling is then followed to produce cluster features.

For 3D visual grounding, ScanRefer~\cite{Chen2020ScanRefer3O} utilizes a localization module to produce a confidence score for each object proposal. The object proposal with the largest confidence score is regarded as the predicted referred object. 

For 3D captioning, Scan2Cap~\cite{Chen2021Scan2CapCD} uses a message-passing graph module to facilitate learning object relation features and uses context-aware attention captioning module~\cite{Anderson2018BottomUpAT} to generate descriptive tokens for each object proposal. Following Scan2Cap~\cite{Chen2021Scan2CapCD}, to jointly measure the quality of the generated descriptions and the detected bounding boxes, we evaluate the descriptions by combining the captioning metric with IoU scores. Thus, the metric is defined as: $m@k$IoU $=\frac{1}{N}\sum_{i=0}^N m_i u_i,$ where $N$ is the total number of object proposals, $m$ is the captioning metric and $u_i \in \{0, 1\}$ is set to 1 if the IoU score for the $i^{th}$ box is greater than $k$. We use the well-established captioning metric of CiDEr~\cite{Vedantam2015CIDErCI}, and employ CiDEr@0.25IoU and CiDEr@0.5IoU where only the prediction whose IoU is larger than 0.25 and 0.5 will be considered.

\subsection{Language Parser}

Our language parser is based on an off-the-shelf rule-based language scene graph parser \footnote{https://github.com/vacancy/scenegraphparser}. It is a python toolkit for parsing sentences (in natural language) into scene graphs based on dependency parsing. We refine the toolkit to better fit the 3D scene data and the corresponding text descriptions by adding additional rules and post-processing. See our code for more details.

\begin{figure}[h]
\begin{center}
\centerline{\includegraphics[width=0.8\columnwidth]{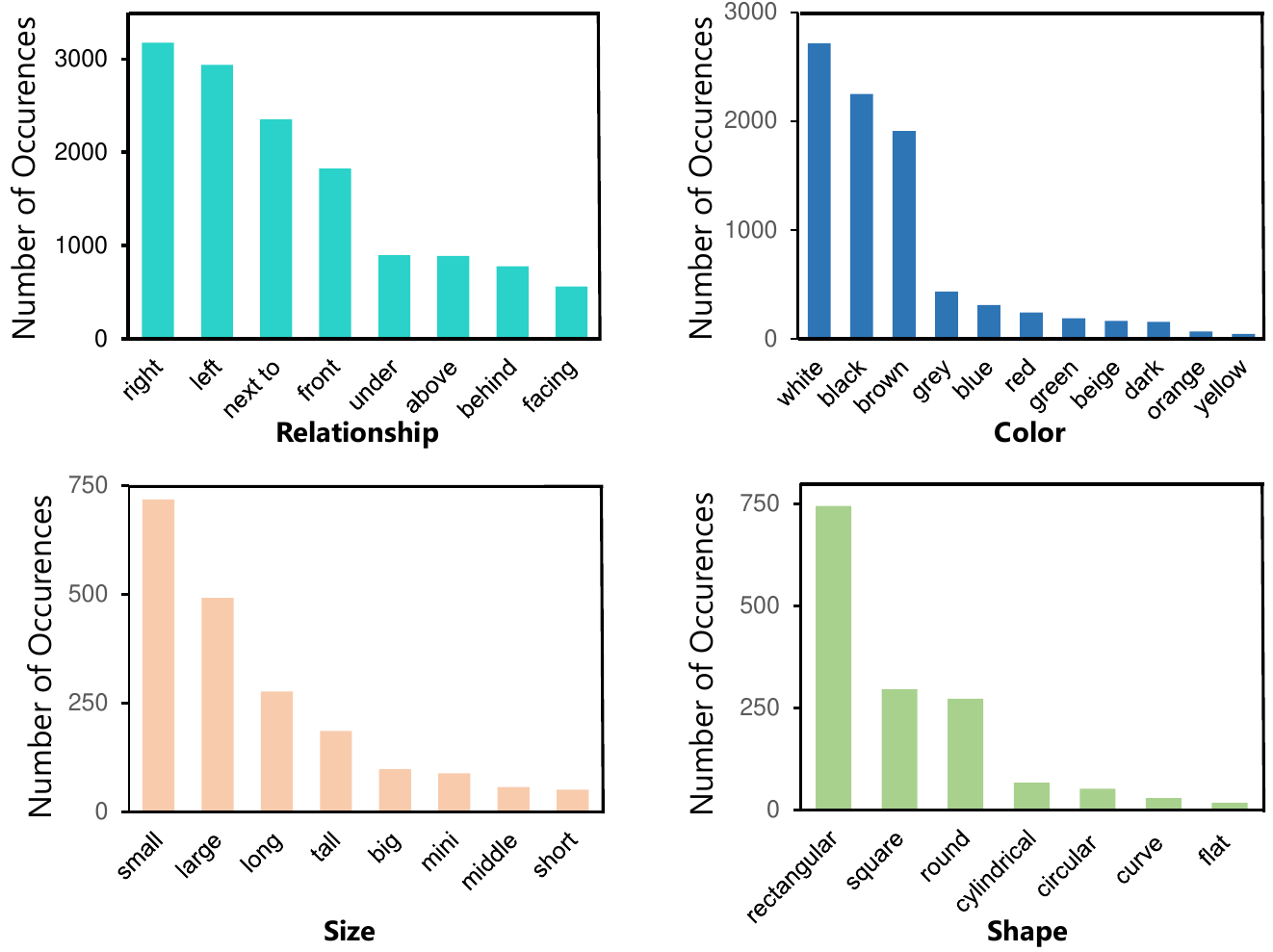}}
\caption{Statistics of the language parsing results.}
\label{statistics}
\end{center}
%\vskip -0.1in
\vspace{-0.6cm}
\end{figure}

Figure~\ref{statistics} shows the data statistics of language parsing results on ScanRefer~\cite{Chen2020ScanRefer3O} dataset. We select 8 relationships and 26 attributes for language-assisted training. The selected attributes can be further divided into color, shape and size. In total, we successfully extract relationships from $12,325$ descriptions and extract attributes from $11,402$ descriptions among the $36,665$ descriptions in ScanRefer’s training set.

\begin{table*}[t]
\footnotesize
  \caption{3D object detection results on ScanNet using limited bounding box annotations.}
  %  Each scene contains an average of 13 bounding boxes.
  \label{data-efficient_appendix}
  \centering
  \begin{tabular}{lcccccccc}
    \toprule
     &  \multicolumn{4}{c}{mAP@0.25}  & \multicolumn{4}{c}{mAP@0.5} \\
    \cmidrule(lr){2-5} \cmidrule(lr){6-9}
    No. of Boxes  & 1 & 2 & 4 & 7    & 1 & 2 & 4 & 7    \\
    \midrule
    VoteNet~\cite{Qi2019DeepHV} &  13.74 & 20.13  &  23.08  & 36.16 & 8.11 & 10.02 &  15.08  & 23.12   \\
    Ours                  &  \textbf{18.55}  &  \textbf{22.33}  & \textbf{30.61}  & \textbf{38.35} & \textbf{9.33}  &  \textbf{11.66}  &  \textbf{17.84}  &  \textbf{23.68}  \\
    \bottomrule
  \end{tabular}
 \vspace{-0.1cm}
\end{table*}

\subsection{Our Modifications to Language Parser}
\label{modification}

The original language parser~\cite{Wu2019UnifiedVE} cannot handle the wide variation in language without updates and adding more rules, especially for some relation descriptions in the languages from ScanRefer dataset. So additional rules are added to the language parser to adapt the model to the 3D scene data and corresponding language descriptions. Here we will describe some critical added rules. More details can be found in the language parser codes.

First, we replace pronouns with the noun phrases of the referred 3D object to solve the problem of pronouns’ reference.

Second, we add some relationship phrases and attribute words to the original language library. However, some added relation phrases are difficult to extract because of their complex forms, such as \textit{“on the left side of”}, \textit{“in front of”}, \textit{“at the top of”}. Thus, we add additional rules to the parser when the sentence root’s head has the dependency of \textit{“prepositional object”}.

Third, we deal with special sentence patterns in ScanRefer’s free-form languages, such as \textit{“there is a white table placed next to the bed.”} In this case, we add rules to the parser when the sentence root’s head has the dependency of \textit{“amod”} or \textit{“advmod”} and its head’s part of speech is \textit{“VERB”}.

After improving the language parser, the relationships of $25,869$ sentences from ScanRefer’s training set are successfully extracted. The total number of sentences with at least one relation phrase in the training set is $31,063$. Thus, $83.28\%$ of the training languages can be successfully parsed. Still, a more powerful language parser based on neural network can be used instead of rule-based parser in future work.

\section{Additional Experiments}

\subsection{Additional Results for Data-efficient 3D Scene Understanding}
\label{effic}

To further explore the effect of language supervision in data-efficient 3D object detection, we show additional results in Table ~\ref{data-efficient_appendix}. We random sample \{1, 2, 4, 7\} bounding boxes per scene to train the detector. The results show a consistent boost in mAP among different conditions.

% \subsection{Additional Results for Visual Grounding} 
% We provide additional visual grounding results on ScanRefer in table \ref{scanrefer}. \textit{Unique} means there is only one object of its class in the scene. \textit{Multiple} means there are distractors of the same class as the referenced object.

% \begin{table*}[t]
% \footnotesize
% \renewcommand\arraystretch{1.0}
%   \centering
%   \caption{Visual grounding results on ScanRefer. \textit{Unique} means there is only one object of its class in the scene. \textit{Multiple} means there are distractors of the same class as the referenced object.}
%   \label{scanrefer}
%   \begin{tabular}{lc|cccccc}
%     \toprule
%     \multirow{2}{*}{Method} & \multirow{2}{*}{Input} & \multicolumn{2}{c}{Unique} & \multicolumn{2}{c}{Multiple} & \multicolumn{2}{c}{Overall} \\
%      & & Acc@0.25 & Acc@0.5   & Acc@0.25 & Acc@0.5    & Acc@0.25 & Acc@0.5   \\     
%     % \midrule
%     % \multicolumn{8}{c}{Results on ScanRefer validation set} \\
%     \midrule
%     ScanRefer & 3D & \textbf{67.37} & 47.86 &  30.77 & 20.50 & 37.87 & 25.81 \\
%     +Ours  & 3D & 66.84 & \textbf{49.80} & \textbf{32.23} & \textbf{22.75} & \textbf{38.95} & \textbf{28.00} \\
%     % \midrule
%     % ScanRefer & 2D+3D & \textbf{79.57} & 56.01 &
%     % 34.95 & 23.52 & 43.61 & 29.83\\
%     % Ours & 2D+3D & 78.72 & \textbf{59.31} & \textbf{35.54} & \textbf{25.07} & \textbf{43.92} & \textbf{31.71}  \\
%     \bottomrule
%   \end{tabular}
%   \vspace{-0.1cm}
% \end{table*}

\subsection{Explorations of Auxiliary Task Design}
\label{exp_aux}

\textbf{How language parsing results affect the 3D feature learning.} In the main paper, we only perform the auxiliary tasks on the 8 most frequent relationships and 26 attributes. Here we analyze how the number of relationship/attribute categories affects the 3D feature learning. The results in Table \ref{num_rel} and \ref{num_attr} indicate that more training categories always bring positive effects. But the introduction of more less frequent categories brings less improvement.

\begin{table}[h]
\footnotesize
  \caption{3D object detection results on ScanNet V2 with different numbers of relationship categories.}
  %  Each scene contains an average of 13 bounding boxes.
  \label{num_rel}
  \centering
  \begin{tabular}{lccc}
    \toprule
    Num of relation &  5 &  8  &  16  \\
    \midrule
    mAP@0.5  & 32.48  &	33.13  & \textbf{33.46}  \\
    \bottomrule
  \end{tabular}
 \vspace{-0.11cm}
\end{table}

\begin{table}[h]
\footnotesize
  \caption{3D object detection results on ScanNet V2 with different numbers of attribute categories.}
  %  Each scene contains an average of 13 bounding boxes.
  \label{num_attr}
  \centering
  \begin{tabular}{lccc}
    \toprule
    Num of attribute &  13 &  26  &  37  \\
    \midrule
    mAP@0.5  & 32.81  &	33.13  & \textbf{33.31}  \\
    \bottomrule
  \end{tabular}
 \vspace{-0.11cm}
\end{table}

\textbf{The loss function design of the relation classification task.} Since the relationship categories are not mutually exclusive, we propose to use binary cross entropy loss for relation classification in the main paper. We also tried traditional cross-entropy classification loss, which also works for language-assisted training but is less effective than the binary cross-entropy loss. The results of 3D object detection are shown in Table \ref{binary}.

\begin{table}[h]
\footnotesize
  \caption{3D object detection results on ScanNet V2 with different relation classification loss designs.}
  %  Each scene contains an average of 13 bounding boxes.
  \label{binary}
  \centering
  \begin{tabular}{lcc}
    \toprule
     &  Binary Cross-entropy & Cross-entropy   \\
    \midrule
    mAP@0.5  &  \textbf{33.13}  &  32.51   \\
    \bottomrule
  \end{tabular}
 \vspace{-0.11cm}
\end{table}

\subsection{Explorations of Multi-modal Feature Fusion Strategy}
\label{exp_fuse}

For simplicity, we fuse the language and 3D features via concatenation to keep the same as ScanRefer and Scan2Cap. This multi-modal feature fusion is proved to be beneficial for both 3D object detection. On ScanNet V2, with VoteNet backbone, feature concatenation would result in an mAP@0.5 of 33.13, compared with 32.14 from the "No fuse" baseline. 

In addition to learn better 3D features, we also hope to emphasize the language encoder’s ability to extract attribute and relationship information from language descriptions. Notice that the language encoder is also used in 3D-language downstream tasks. Thus, fusing language embeddings and 3D features to perform the auxiliary tasks helps the language encoder to encode abundant attribute and scene context information, and provides more clues for 3D visual grounding and captioning. The results of visual grounding task with ScanRefer backbone are shown in Table \ref{vg_fuse}, where “No fuse” represents only using 3D features to perform relationship classification and attribute classification tasks. Experimental results confirm that the fusion process benefits the visual grounding task.

\begin{table}[h]
\footnotesize
  \caption{Visual grounding results on ScanRefer.}
  %  Each scene contains an average of 13 bounding boxes.
  \label{vg_fuse}
  \centering
  \begin{tabular}{lcc}
    \toprule
     &  ACC@0.25 &  ACC@0.5  \\
    \midrule
    No fuse   & 37.88  & 26.04  \\
    Concatenation  &   \textbf{38.95}  & \textbf{28.00} \\
    \bottomrule
  \end{tabular}
 \vspace{-0.11cm}
\end{table}

Besides applying feature fusion with concatenation, another option is to utilize a QKV attention module to guide the multi-modal interaction process better. Specifically, language features serve as query vectors and 3D features as key/value vectors to compute the final multi-modal features. This module can be easily incorporated into our framework and our experiments in Table \ref{attention} show that this more complex fusion strategy could work better than concatenation. “No fuse” means that only 3D features are used to perform auxiliary tasks.

\begin{table}[h]
\footnotesize
  \caption{3D object detection results on ScanNet V2 with different fusion strategies.}
  %  Each scene contains an average of 13 bounding boxes.
  \label{attention}
  \centering
  \begin{tabular}{lcc}
    \toprule
     &  mAP@0.25 &  mAP@0.5  \\
    \midrule
    Concatenation   & 51.91  & 33.13  \\
    Attention  &   \textbf{52.58}  & \textbf{33.41} \\
    \bottomrule
  \end{tabular}
 \vspace{-0.11cm}
\end{table}

\subsection{Baseline Method of Language-assisted Training}
\label{lang}

In this section, we explore other baseline methods that also consider language information during the training of 3D scene understanding network. Since the 3D detector is trained end-to-end in the 3D visual grounding task, the supervision of languages also affects the training of object detection. Thus, we take the 3D detector trained with the visual grounding objective as a basic language-assisted trained model, which is denoted as “Visual Grounding” in Table ~\ref{baseline}. The results of mAP@0.5 show that this basic language-assisted training method has little effect on object detection performance, indicating that parsing the languages and introducing the cross-modal knowledge at object-level is necessary for language-assisted 3D feature learning.

\begin{table}[h]
\footnotesize
  \caption{3D object detection results on ScanNet.}
  %  Each scene contains an average of 13 bounding boxes.
  \label{baseline}
  \centering
  \begin{tabular}{lc}
    \toprule
    Method &  mAP \\
    \midrule
    VoteNet~\cite{Qi2019DeepHV}   & 31.67 \\
    Visual Grounding  &  31.49  \\
    Ours  &   \textbf{33.13} \\
    \bottomrule
  \end{tabular}
 \vspace{-0.11cm}
\end{table}

\subsection{Language-assisted 3D Feature Learning with More Informative Languages}
\label{det}

To further explore the potential of language-assisted training, we train the auxiliary tasks with those languages from which at least one relationship or attribute is extracted. In this way, auxiliary tasks affect the 3D scene understanding model in every iteration. In total, $19,364$ descriptions among the $36,665$ descriptions in ScanRefer’s training set are selected in this paradigm. Table ~\ref{select} shows that our method outperforms the VoteNet~\cite{Qi2019DeepHV} baseline significantly, indicating that training the model with more informative languages is more beneficial for 3D feature learning.

\begin{table}[h]
\footnotesize
  \caption{3D object detection results with selected languages' assistance.}
  %  Each scene contains an average of 13 bounding boxes.
  \label{select}
  \centering
  \begin{tabular}{lcc}
    \toprule
    Method &  mAP@0.25 &  mAP@0.5  \\
    \midrule
    VoteNet~\cite{Qi2019DeepHV}   & 46.48  & 27.25  \\
    Ours  &   \textbf{50.62}  & \textbf{30.51} \\
    \bottomrule
  \end{tabular}
 \vspace{-0.11cm}
\end{table}

\subsection{Comparison with Pre-training Methods}
\label{com_pretrain}

Existing pre-training methods for 3D tasks include pre-training on a 3D visual dataset, unsupervised/self-supervised pre-training~\cite{Xie2020PointContrastUP}, and multi-modal pre-training with the help of 2D images and 2D encoders~\cite{Afham2022CrossPointSC}. To the best of our knowledge, there are no 3D-language pre-training methods so far.

Our method differs from the existing pre-training methods in several ways:

First, we explore the supervision of another modality, namely the free-form language, to improve 3D scene understanding, which is not involved in previous studies.

Second, the target of the existing pre-training method and our language-assisted training are different. Specifically, we aim to guide 3D feature learning toward important attributes and scene context information. In contrast, traditional pre-training and 3D scene understanding methods only focus on the semantic and localization information of the objects. Thus, our training objectives can be complementary to existing pre-training methods.

Third, the 3D features learned with language assistance contain abundant attribute and scene context information, which provides more clues for 3D visual grounding and captioning tasks than the traditional 3D encoder.

Our method can also combine with the pre-training methods and further improve the results:

\textbf{Improving supervised pre-training models.} Here we show the instance segmentation results on ScanNet V2 with SoftGroup~\cite{Vu2022SoftGroupF3}. We use the checkpoint of HAIS~\cite{Chen2021HierarchicalAF} to initialize the U-Net backbone and the MLP in semantic/offset branches. The results in Table \ref{softgroup_pretrain} indicate that our method can further improve 3D feature learning based on the pre-trained model.

\begin{table}[h]
\footnotesize
  \caption{3D instance segmentation results on ScanNet V2.}
  %  Each scene contains an average of 13 bounding boxes.
  \label{softgroup_pretrain}
  \centering
  \begin{tabular}{lcc}
    \toprule
     &  mAP@0.25 &  mAP@0.5  \\
    \midrule
    SoftGroup~\cite{Vu2022SoftGroupF3}   & 78.4  & 66.0  \\
    Ours  &   \textbf{79.5}  & \textbf{66.8} \\
    \bottomrule
  \end{tabular}
 \vspace{-0.11cm}
\end{table}

\section{Visualization}

\begin{figure}[h]
\begin{center}
\centerline{\includegraphics[width=0.9\columnwidth]{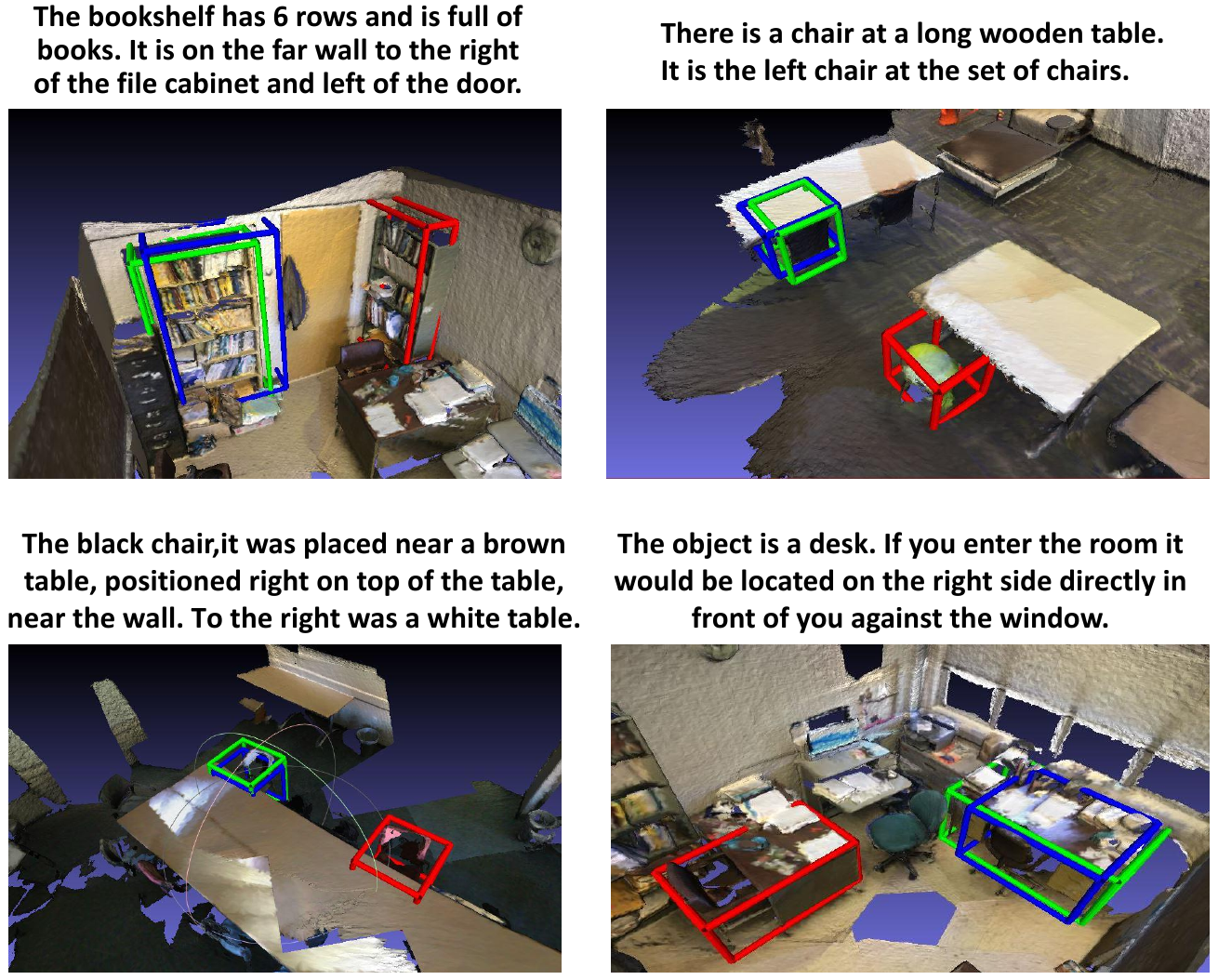}}
\caption{More visualization results of visual grounding. The ground-truth boxes are displayed in blue. The predicted boxes of our method and the baseline are displayed in green and red.}
\label{visual_vg}
\end{center}
%\vskip -0.1in
\vspace{-0.6cm}
\end{figure}

We provide additional visualization results of visual grounding and object detection. Figure ~\ref{visual_vg} shows that our method can better utilize the relationships and attributes information in the language to localize the target objects. In the first picture, there are two bookshelves in the scene, and the target of the language input is the left one (in blue). Two relation triplets can be extracted from the language description, namely, (bookshelf, right of, cabinet) and (bookshelf, left of, door). Our model better preserves this relation information and finds the correct bookshelf (in green). In contrast, the baseline model fails to infer from relational information and predicts the wrong results (in red). In the second picture, there are two chairs in the scene, and the target one is on the left (in blue). In the language descriptions, the attribute word “long” used to describe the table is an essential basis for finding the target object. Our model better utilizes this clue and predicts the right location, while the baseline method fails to distinguish between the two tables. 

\begin{figure}[h]
\begin{center}
\centerline{\includegraphics[width=0.8\columnwidth]{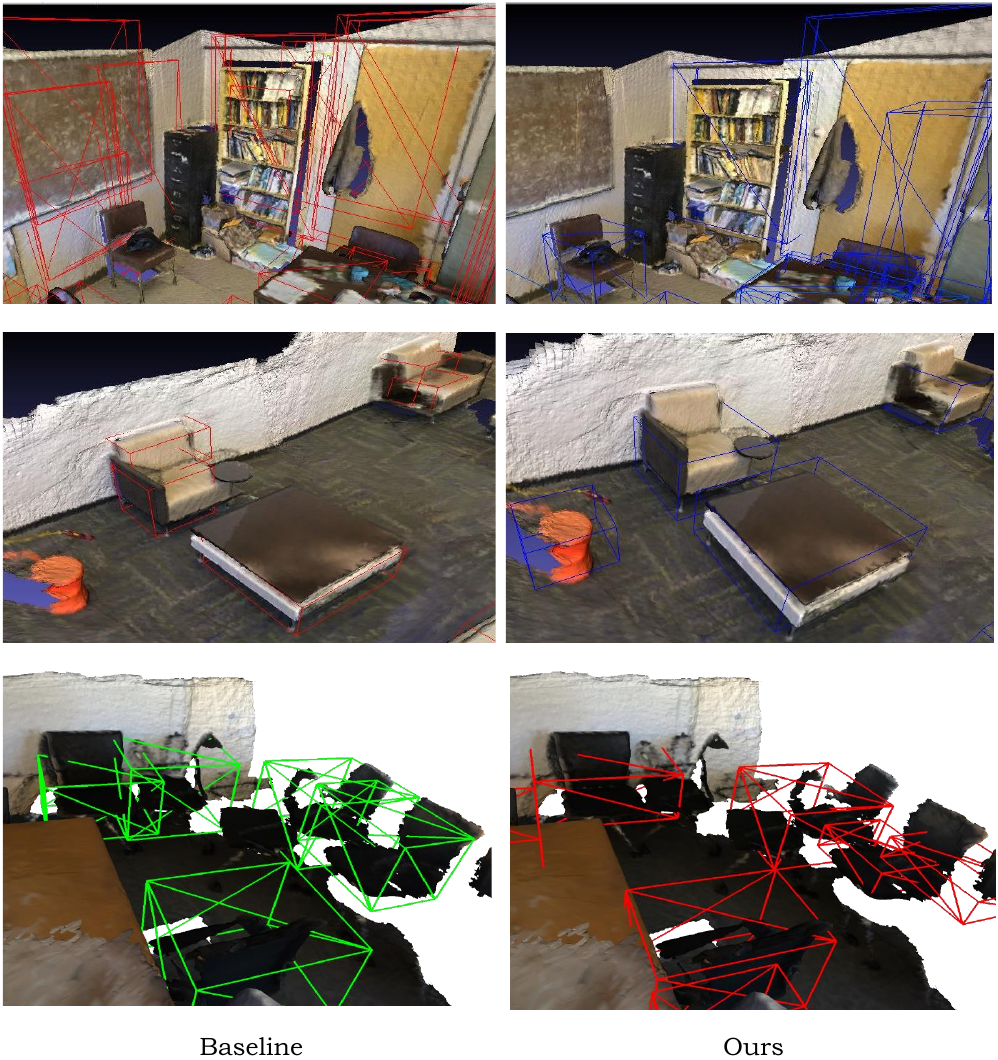}}
\caption{More visualization results of object detection. The ground-truth boxes are displayed in blue. The predicted boxes of our method and the baseline are displayed in green and red.}
\label{visual_det}
\end{center}
%\vskip -0.1in
\vspace{-0.6cm}
\end{figure}

Figure ~\ref{visual_det} shows that our method can produce more compact boxes for large objects such as bookshelves and doors, and produce more accurate boxes for the L-shaped sofas.

\end{document}